# Title:
# Quantum-inspired canonical correlation analysis for exponentially large dimensional data

**Authors:**
**Naoko Koide-Majima**[a,b]†**, Kei Majima**[c,d]†

**Affiliations:**
[a] AI Science Research and Development Promotion Center, National Institute of Information and Communications Technology, Osaka 565-0871, Japan
[b] Graduate School of Frontier Biosciences, Osaka University, Osaka 565-0871, Japan
[c] Graduate School of Informatics, Kyoto University, Kyoto 606-8501, Japan
[d] Institute for Quantum Life Science, National Institutes for Quantum and Radiological Science and Technology, Chiba, 263-8555, Japan
† These authors contributed equally to this work.

**Corresponding author:**
Kei Majima, Ph.D.
Institute for Quantum Life Science,
National Institutes for Quantum and Radiological Science and Technology,
Chiba, 263-8555, Japan
Phone: +81-43-251-2111
E-mail: majima.kei@qst.go.jp

**Declaration of interest:**
None.

# Abstract

Canonical correlation analysis (CCA) serves to identify statistical dependencies between pairs of multivariate data. However, its application to high-dimensional data is limited due to considerable computational complexity. As an alternative to the conventional CCA approach that requires polynomial computational time, we propose an algorithm that approximates CCA using quantum-inspired computations with computational time proportional to the logarithm of the input dimensionality. The computational efficiency and performance of the proposed quantum-inspired CCA (qiCCA) algorithm are experimentally evaluated on synthetic and real datasets. Furthermore, the fast computation provided by qiCCA allows directly applying CCA even after nonlinearly mapping raw input data into high-dimensional spaces. The conducted experiments demonstrate that, as a result of mapping raw input data into the high-dimensional spaces with the use of second-order monomials, qiCCA extracts more correlations compared with the linear CCA and achieves comparable performance with state-of-the-art nonlinear variants of CCA on several datasets. These results confirm the appropriateness of the proposed qiCCA and the high potential of quantum-inspired computations in analyzing high-dimensional data.

# 1. Introduction

Canonical correlation analysis (CCA) is intended to identify statistical dependencies between pairs of multivariate data (Hotelling, 1936). CCA aims at searching a pair of linear projections that map a given pair of multivariate data into a common lower-dimensional space. In the case of appropriate application, the projected data in the lower-dimensional space are maximally correlated. Considering that CCA and its variants allow identifying the common sources of variability in paired data, they are widely applied to investigate associations between data from two different modalities/views. For example, in neuroscience, CCA was employed to reconstruct or to identify viewed images from brain activity measured through functional magnetic resonance imaging (fMRI)(Fujiwara et al., 2013) or electroencephalography (Cao et al., 2015; Dmochowski et al., 2018; Nakanishi et al., 2015). Similarly, in another recent neuroimaging study, CCA was applied to extract the components that could explain human fMRI data based on behavioral data (Koide-Majima et al., 2020). CCA was also utilized to predict personality traits (Smith et al., 2015) and the risk of mental disorders (Yahata et al., 2016) from human fMRI data. In bioinformatics, CCA was applied to find a set of genes in which expressions would be correlated with copy number changes in the DNA (Waaijenborg et al., 2008; Witten et al., 2009) and with the concentration changes of hepatic fatty acids (González et al., 2009; Yoshida et al., 2017). CCA was also utilized to infer protein networks from genomic data (Yamanishi et al., 2004). Moreover, CCA was implemented as a method for cross-modal retrieval using image, text, and video data (Gong et al., 2014; Rasiwasia et al., 2010; K. Wang et al., 2016).

Extended versions of CCA are used to capture nonlinear dependencies between pairs of data. In kernel CCA (Akaho, 2001; Bach & Jordan, 2002; Hardoon et al., 2004; Melzer et al., 2001), input data in each view are nonlinearly mapped into a high-dimensional space, and the mapped data are processed using the linear CCA. The use of kernels in kernel CCA allows avoiding explicit computation of the nonlinear mappings. Similarly, deep CCA (Andrew et al., 2013; Karami & Schuurmans, 2020; W. Wang et al., 2015, 2017) is aimed to construct nonlinear mappings using a pair of neural networks, which are trained to maximally correlate their outputs.

Although CCA is a widely used technique, its application to high-dimensional data (*i.e.,* data with a large number of input features) is limited by its considerable computational complexity. The CCA algorithm employs a singular value decomposition (SVD) as a subroutine (Hsu et al., 2012; Press, 2011), and this step requires $O\left(\min\left(D_1^2 D_2, D_1 D_2^2\right)\right)$ computational time, where $D_1$ and $D_2$ denote the numbers of input features in paired input datasets (*i.e.,* the numbers of columns in the paired input data matrices). When the numbers of features are large, the matrix decomposition is unfeasibly time-consuming. The CCA algorithm based on a stochastic gradient descent method has been introduced to mitigate this problem (Ma et al., 2015); however, it still requires $O(D_1 + D_2)$ computational time per iteration. Furthermore, the actual computational time of this method considerably depends on the stopping criterion. Once the kernel matrices are computed, kernel CCA can be executed in constant time regardless of the dimensionalities of the data in the original spaces (Akaho, 2001; Bach & Jordan, 2002; Hardoon et al., 2004; Melzer et al., 2001). However, the standard kernel CCA algorithm requires solving an $(N \times N)$-matrix eigenvalue problem, where $N$ is the number of samples, which is often intractable for big (many-sample) data. Another algorithm to approximate kernel CCA has been developed based on a technique referred to as random Fourier features (Uurtio et al., 2019; W. Wang & Livescu, 2016), which requires at least $O(D_1 + D_2)$ computational time. Thus, an algorithm that approximates CCA with ultra-low computational time has remained elusive.

Recently, fast machine learning algorithms have been developed by "dequantizing" quantum machine learning algorithms designed to run on quantum computers. Inspired by the quantum algorithm for recommendation systems (Kerenidis & Prakash, 2016), Tang (2019) proposed a classical randomized algorithm to approximate right singular vectors of a given $N \times D$ matrix in $O(\log(ND))$ computational time (in the case of fixing the parameter controlling approximation performance; see Subsection 3.3). In the present study, this algorithm is referred to as quantum-inspired singular value decomposition (qiSVD). Adopting techniques from randomized linear algebra (Frieze et al., 1998), qiSVD performs SVD on a small matrix whose rows and columns have been sampled from a given matrix, and transforms the SVD results to approximate the right singular vectors of the original given matrix. Subsequent algorithms based on this work have been developed to approximate principal component analysis (Tang, 2018), linear regression (Chia et al., 2018; Gilyén et al., 2018), non-negative matrix factorization (Chen et al., 2019) and support vector machine (Ding et al., 2019). These approximation algorithms achieve exponentially faster speed than their known exact-algorithm counterparts.

By introducing qiSVD into the CCA algorithm, we develop an algorithm that approximates CCA with computational time proportional to the logarithm of the input dimensionalities. In the original qiSVD algorithm proposed by Tang (2019), the approximate right singular vectors are not exactly orthonormal, which significantly degrades the performance of CCA. To resolve this problem, we modify the qiSVD algorithm to return a set of orthonormal vectors. Also, whereas previous qiSVD studies identified the theoretical upper bounds of the approximation error in qiSVD (Tang, 2019), or evaluated the approximation performance on artificial data (Arrazola et al., 2019), we demonstrate the performances of qiSVD and qiCCA on five real datasets from various fields, including image, text, and biometric data.

Furthermore, the fast computation of qiCCA allows us to directly apply CCA even after nonlinearly mapping raw input data into high-dimensional spaces. To demonstrate this capability, we computed the products of the input features for all possible feature pairs (Majima et al., 2014), and applied qiCCA to the resulting high-dimensional data. This procedure increases the number of input dimensions in each view from $D$ in the original space to $D(D-1)/2$. Processing a dataset of this size is often computationally infeasible when using the conventional CCA. Although kernel CCA with nonlinear kernel functions implicitly addresses such high-dimensional spaces, it becomes computationally infeasible in the case of data with many samples. Using the above procedure, we demonstrate that qiCCA extracts more correlations than the linear CCA and performs comparably to state-of-the-art nonlinear variants of CCA. The combination of qiCCA and nonlinear mapping of numerous basis functions might provide a new technique for finding nonlinear correlations.

The main contributions of this paper are as follows.

- Previous studies focused on the theoretical upper bounds of the approximation error of qiSVD. Instead, we examine the approximation performance of qiSVD on five real datasets from various fields, including image, text, and biometric data.

- The original qiSVD algorithm proposed by Tang et al. (2019) returns approximate right singular vectors that are not exactly orthonormal. After the introduced modification, the qiSVD algorithm developed as a result of this research returns a set of orthonormal vectors.

- The modified qiSVD algorithm is embedded into an algorithm that approximates CCA with computational time proportional to the logarithm of the input dimensionalities.

- Combined with a large number of polynomial basis functions, we demonstrate that the fast computation of our proposed algorithm provides a new technique for revealing nonlinear correlations.

The remainder of this paper is organized as follows. Section 2 provides a brief overview on the related work. The CCA, qiSVD, and qiCCA algorithms are introduced in Section 3. Section 4 presents the experimental results. Discussion and concluding remarks are given in Sections 5 and 6, respectively.

# 2. Related work

## 2.1. Quantum-inspired algorithms

Since Tang (2019) released the qiSVD algorithm for low-rank matrix approximations, researchers have proposed several quantum-inspired algorithms. For example, approximation algorithms for principal component analysis (Tang, 2018), linear regression (Gilyén et al., 2018; Chia et al., 2018), non-negative matrix factorization (Chen et al., 2019), and support vector machine (Ding et al., 2019) have been developed. All these studies adopted the qiSVD algorithm proposed by Tang (2019). The resulting time complexities are exponentially lower than their existing counterparts. Similarly to those previous studies, we incorporate qiSVD into an algorithm that approximates CCA with an exponentially lower computational time than the known exact CCA algorithm. It should be noted that the original qiSVD algorithm proposed by Tang (2019) returns approximate right singular vectors that are not exactly orthonormal. We demonstrate that this anomaly significantly degrades the performance of CCA. Therefore, we modify the qiSVD algorithm to return a set of orthonormal vectors.

We consider that qiSVD can be regarded as a variant of randomized SVD (Halko et al., 2011), which transforms an input matrix into a small matrix, applies SVD to the small matrix, and back-projects the SVD results to approximate the SVD results of the original matrix. This framework is also employed in qiSVD. For an $I \times J$ input matrix, constructing a small matrix requires $O(IJ)$ computational time in randomized SVD and $O(\log(IJ))$ computational time in qiSVD. It should be noted that, because our main purpose is approximating CCA with a computational time proportional to the logarithm of the input dimensionality, a comparison of qiSVD with randomized SVD is outside the scope of this study. However, investigating and comparing the advantages and disadvantages of qiSVD and randomized SVD remains an important challenge, which may provide important insights into the properties of qiSVD.

## 2.2. Other approaches for scalable CCA

Researchers reported several attempts to reduce the computational time of CCA. As mentioned above, the standard CCA algorithm based on SVD requires $O(\min(D_1^2 D_2, D_1 D_2^2))$ computational time, where $D_1$ and $D_2$ are the input dimensionalities of the given paired data. To resolve the problem of such time-consuming matrix decomposition, Ma et al., (2015) approximated the CCA using the iterative gradient method, which reduces the time complexity to $O(D_1 + D_2)$. As an alternative approach, once the kernel matrices are computed, kernel CCA can operate in constant time regardless of the input dimensionalities (Akaho, 2001; Bach & Jordan, 2002; Hardoon et al., 2004; Melzer et al., 2001). Nevertheless, the computational time of kernel CCA is proportional to the cube of the number of input samples, and easily becomes infeasible for big data. To address this issue, researchers have developed approximate kernel CCA algorithms based on random Fourier features (Uurtio et al., 2019; W. Wang & Livescu, 2016), which consume $O(D_1 + D_2)$ computational time.

# 3. Algorithm

This section is organized as follows. We first explain the mathematical notations, and then, introduce the conventional CCA algorithm which uses an SVD subroutine. In the following subsections, we explain qiSVD and the data structure that enables logarithmic-time qiSVD. Finally, we explain the qiCCA algorithm. A Python implementation of qiCCA is available at our GitHub repository [https://github.com/nkmjm/qiML].

## 3.1. Notations

We denote the set of $I \times J$ real matrices by $\mathbb{R}^{I \times J}$. For each matrix $\mathbf{A} \in \mathbb{R}^{I \times J}$, we denote the $(i,j)$-entry by $\mathbf{A}(i,j)$, the $i$-th row by $\mathbf{A}(i,:)$, and the $j$-th column by $\mathbf{A}(:,j)$. The $I \times K$ matrix composed of the first $K$ columns of $\mathbf{A}$ is denoted by $\mathbf{A}(:,1:K)$, and the $K \times J$ matrix composed of the first $K$ rows of $\mathbf{A}$ is denoted by $\mathbf{A}(1:K,:)$. The transpose, inverse, and Frobenius norm of $\mathbf{A}$ are denoted by $\mathbf{A}^\mathrm{T}$, $\mathbf{A}^{-1}$, and $\|\mathbf{A}\|_F$, respectively. Similarly, we denote the set of $N$-dimensional real column vectors by $\mathbb{R}^N$. The $n$-th element of a vector $\mathbf{v} \in \mathbb{R}^N$ is denoted by $\mathbf{v}(n)$. The transpose and Euclidean norm of $\mathbf{v}$ are denoted by $\mathbf{v}^\mathrm{T}$ and $\|\mathbf{v}\|$, respectively.

## 3.2. Canonical correlation analysis

We introduce the CCA algorithm based on the SVD subroutine. We assume a pair of data matrices, $\mathbf{X} \in \mathbb{R}^{N \times D_1}$ and $\mathbf{Y} \in \mathbb{R}^{N \times D_2}$, where $N$ is the number of samples and $D_1$ and $D_2$ are the numbers of input dimensions (*i.e.,* input features) in views 1 and 2, respectively. In CCA, the first pair of linear weight vectors $\widehat{\mathbf{w}}_X^{(1)} \in \mathbb{R}^{D_1}$ and $\widehat{\mathbf{w}}_Y^{(1)} \in \mathbb{R}^{D_2}$ is defined by maximizing the following objective function:

$$\widehat{\mathbf{w}}_X^{(1)}, \widehat{\mathbf{w}}_Y^{(1)} = \underset{\mathbf{w}_X^{(1)}, \mathbf{w}_Y^{(1)}}{\operatorname{argmax}} \operatorname{corr}\left[\mathbf{X}\mathbf{w}_X^{(1)}, \mathbf{Y}\mathbf{w}_Y^{(1)}\right], \qquad (1)$$

where the $\operatorname{corr}[\cdot,\cdot]$ function returns the Pearson correlation coefficient between two input column vectors. The objective function can be rewritten as

$$\widehat{\mathbf{w}}_X^{(1)}, \widehat{\mathbf{w}}_Y^{(1)} = \underset{\mathbf{w}_X^{(1)}, \mathbf{w}_Y^{(1)}}{\operatorname{argmax}} \frac{\mathbf{w}_X^{(1)\mathrm{T}} \mathbf{\Sigma}_{XY} \mathbf{w}_Y^{(1)}}{\sqrt{\mathbf{w}_X^{(1)\mathrm{T}} \mathbf{\Sigma}_{XX} \mathbf{w}_X^{(1)}} \sqrt{\mathbf{w}_Y^{(1)\mathrm{T}} \mathbf{\Sigma}_{YY} \mathbf{w}_Y^{(1)}}}, \qquad (2)$$

where $\mathbf{\Sigma}_{XX}$ and $\mathbf{\Sigma}_{YY}$ are the sample covariance matrices derived from $\mathbf{X}$ and $\mathbf{Y}$, respectively, and $\mathbf{\Sigma}_{XY}$ is the sample cross-covariance matrix. As this objective function is invariant with respect to scaling of $\mathbf{w}_X^{(1)}$ and $\mathbf{w}_Y^{(1)}$, we introduce the following constraints:

$$\mathbf{w}_X^{(1)^\mathrm{T}} \mathbf{\Sigma}_{XX} \mathbf{w}_X^{(1)} = 1, \quad \mathbf{w}_Y^{(1)^\mathrm{T}} \mathbf{\Sigma}_{YY} \mathbf{w}_Y^{(1)} = 1. \tag{3}$$

The linear weight vectors are therefore defined as follows:

$$\widehat{\mathbf{w}}_X^{(1)}, \widehat{\mathbf{w}}_Y^{(1)} = \underset{\mathbf{w}_X^{(1)^\mathrm{T}} \mathbf{\Sigma}_{XX} \mathbf{w}_X^{(1)} = 1, \ \mathbf{w}_Y^{(1)^\mathrm{T}} \mathbf{\Sigma}_{YY} \mathbf{w}_Y^{(1)} = 1}{\mathrm{argmax}} \mathbf{w}_X^{(1)^\mathrm{T}} \mathbf{\Sigma}_{XY} \mathbf{w}_Y^{(1)}. \tag{4}$$

After deriving the first pair of linear weight vectors using Eq. (4), the second pair is obtained as

$$\widehat{\mathbf{w}}_X^{(2)}, \widehat{\mathbf{w}}_Y^{(2)} = \underset{\mathbf{w}_X^{(2)^\mathrm{T}} \mathbf{\Sigma}_{XX} \mathbf{w}_X^{(2)} = 1, \ \mathbf{w}_Y^{(2)^\mathrm{T}} \mathbf{\Sigma}_{YY} \mathbf{w}_Y^{(2)} = 1}{\mathrm{argmax}} \mathbf{w}_X^{(2)^\mathrm{T}} \mathbf{\Sigma}_{XY} \mathbf{w}_Y^{(2)}, \tag{5}$$

subject to

$$\widehat{\mathbf{w}}_X^{(1)^\mathrm{T}} \mathbf{\Sigma}_{XX} \mathbf{w}_X^{(2)} = 0, \quad \widehat{\mathbf{w}}_Y^{(1)^\mathrm{T}} \mathbf{\Sigma}_{YY} \mathbf{w}_Y^{(2)} = 0. \tag{6}$$

Under the above constraints, the first and second canonical variates are uncorrelated with each other. The subsequent pairs of linear weight vectors are similarly obtained as

$$\widehat{\mathbf{w}}_X^{(k)}, \widehat{\mathbf{w}}_Y^{(k)} = \underset{\mathbf{w}_X^{(k)^\mathrm{T}} \mathbf{\Sigma}_{XX} \mathbf{w}_X^{(k)} = 1, \ \mathbf{w}_Y^{(k)^\mathrm{T}} \mathbf{\Sigma}_{YY} \mathbf{w}_Y^{(k)} = 1}{\mathrm{argmax}} \mathbf{w}_X^{(k)^\mathrm{T}} \mathbf{\Sigma}_{XY} \mathbf{w}_Y^{(k)} \quad (k = 1, \cdots, K), \tag{7}$$

subject to

$$\widehat{\mathbf{w}}_X^{(k_1)^\mathrm{T}} \mathbf{\Sigma}_{XX} \mathbf{w}_X^{(k_2)} = 0, \quad \widehat{\mathbf{w}}_Y^{(k_1)^\mathrm{T}} \mathbf{\Sigma}_{YY} \mathbf{w}_Y^{(k_2)} = 0 \quad (0 < k_1 < k_2 \leq K). \tag{8}$$

Here, $K$ is the manually specified number of the pairs to be computed.

The above weight vectors can be computed by various methods. Here, we follow the SVD-based approach (Chu et al., 2013; Hsu et al., 2012; Press, 2011). The SVDs of input matrices $\mathbf{X}$ and $\mathbf{Y}$ are respectively computed as follows:

$$\mathbf{X} = \mathbf{U}_1 \mathbf{\Sigma}_1 \mathbf{V}_1^T, \qquad \mathbf{Y} = \mathbf{U}_2 \mathbf{\Sigma}_2 \mathbf{V}_2^T, \qquad (9)$$

where $\mathbf{U}_1 \in \mathbb{R}^{N \times \text{rank}(\mathbf{X})}$ and $\mathbf{U}_2 \in \mathbb{R}^{N \times \text{rank}(\mathbf{Y})}$ are the matrices composed of the left singular vectors of $\mathbf{X}$ and $\mathbf{Y}$, respectively; $\mathbf{\Sigma}_1 \in \mathbb{R}^{\text{rank}(\mathbf{X}) \times \text{rank}(\mathbf{X})}$ and $\mathbf{\Sigma}_2 \in \mathbb{R}^{\text{rank}(\mathbf{Y}) \times \text{rank}(\mathbf{Y})}$ are the diagonal matrices whose diagonal entries are the singular values of $\mathbf{X}$ and $\mathbf{Y}$, respectively; $\mathbf{V}_1 \in \mathbb{R}^{D_1 \times \text{rank}(\mathbf{X})}$ and $\mathbf{V}_2 \in \mathbb{R}^{D_2 \times \text{rank}(\mathbf{Y})}$ are the matrices composed of the right singular vectors of $\mathbf{X}$ and $\mathbf{Y}$, respectively. We also apply SVD to $\mathbf{U}_1^T \mathbf{U}_2$ as follows:

$$\mathbf{U}_1^T \mathbf{U}_2 = \mathbf{U}_3 \mathbf{\Sigma}_3 \mathbf{V}_3^T. \qquad (10)$$

The $k$-th pair of the linear weight vectors is obtained as

$$\widehat{\mathbf{w}}_X^{(k)} = \mathbf{V}_1 \mathbf{\Sigma}_1^{-1} \mathbf{U}_3(:,k), \qquad \widehat{\mathbf{w}}_Y^{(k)} = \mathbf{V}_2 \mathbf{\Sigma}_2^{-1} \mathbf{V}_3(:,k). \qquad (11)$$

For the convenience of notation, we denote the linear weight vectors as follows:

$$\mathbf{W}_X = \left[ \widehat{\mathbf{w}}_X^{(1)}, \cdots, \widehat{\mathbf{w}}_X^{(K)} \right], \qquad \mathbf{W}_Y = \left[ \widehat{\mathbf{w}}_Y^{(1)}, \cdots, \widehat{\mathbf{w}}_Y^{(K)} \right]. \qquad (12)$$

These matrices can be obtained as below:

$$\mathbf{W}_X = \mathbf{V}_1 \mathbf{\Sigma}_1^{-1} \mathbf{U}_3(:,1:K), \qquad \mathbf{W}_Y = \mathbf{V}_2 \mathbf{\Sigma}_2^{-1} \mathbf{V}_3(:,1:K). \qquad (13)$$

The SVD-based CCA algorithm is summarized as Algorithm 1.

> **Algorithm 1: CCA**
>
> **Input:** Input matrices $\mathbf{X} \in \mathbb{R}^{N \times D_1}$, $\mathbf{Y} \in \mathbb{R}^{N \times D_2}$, the number of components $K$
>
> **Output:** $K$ pairs of linear weight vectors $\mathbf{W}_X \in \mathbb{R}^{D_1 \times K}$ and $\mathbf{W}_Y \in \mathbb{R}^{D_2 \times K}$
>
> 1: $[\mathbf{U}_1, \mathbf{\Sigma}_1, \mathbf{V}_1] \leftarrow \mathrm{SVD}(\mathbf{X})$
> 2: $[\mathbf{U}_2, \mathbf{\Sigma}_2, \mathbf{V}_2] \leftarrow \mathrm{SVD}(\mathbf{Y})$
> 3: $[\mathbf{U}_3, \mathbf{\Sigma}_3, \mathbf{V}_3] \leftarrow \mathrm{SVD}(\mathbf{U}_1^\mathrm{T} \mathbf{U}_2)$
> 4: $\mathbf{W}_X \leftarrow \mathbf{V}_1 \mathbf{\Sigma}_1^{-1} \mathbf{U}_3(:,1:K)$
> 5: $\mathbf{W}_Y \leftarrow \mathbf{V}_2 \mathbf{\Sigma}_2^{-1} \mathbf{V}_3(:,1:K)$

Using the above weight matrices, the canonical variates of $\mathbf{X}$ and $\mathbf{Y}$ are determined as follows:

$$\mathbf{C}_X = \mathbf{X} \mathbf{W}_X = \mathbf{U}_1 \mathbf{U}_3(:,1:K),$$
$$\mathbf{C}_Y = \mathbf{Y} \mathbf{W}_Y = \mathbf{U}_2 \mathbf{V}_3(:,1:K). \tag{14}$$

### 3.3. Quantum-inspired singular value decomposition

In this subsection, we discuss the qiSVD algorithm proposed by Tang (2019). For a given matrix $\mathbf{X} \in \mathbb{R}^{I \times J}$, the standard SVD algorithm provides a low-rank ($K$-rank) approximation $\mathbf{U}\mathbf{\Sigma}\mathbf{V}^\mathrm{T}$. Here, $\mathbf{\Sigma} \in \mathbb{R}^{K \times K}$ is the diagonal matrix whose $(k,k)$-th entry is the $k$-th largest singular value, and $\mathbf{U} \in \mathbb{R}^{I \times K}$ and $\mathbf{V} \in \mathbb{R}^{J \times K}$ are the matrices whose columns are the left and right singular vectors corresponding to the top $K$ singular values, respectively. Then, qiSVD computes a so called "description" of $\mathbf{V}$ (Frieze et al., 1998; Tang, 2019), defined as the set of a matrix $\mathbf{S} \in \mathbb{R}^{P \times J}$ and vectors $\mathbf{u}_k \in \mathbb{R}^P (k = 1,2,\cdots,K)$ that approximate $\mathbf{V}(:,k)$ by $\mathbf{S}^\mathrm{T}\mathbf{u}_k$. Here, $P$ is a parameter that regulates the trade-off between the computational time and approximation performance. Once a description of $\mathbf{V}$ is obtained, it is possible to access any entry of the approximation of $\mathbf{V}$ in constant time, regardless of the size of the input matrix. The qiSVD procedure for a given matrix $\mathbf{X} \in \mathbb{R}^{I \times J}$ is described below.

The qiSVD process results in constructing a small $P \times P$ matrix $\mathbf{W}$ by selecting $P$ rows and $P$ columns from the input matrix $\mathbf{X}$. This small matrix is processed by the conventional SVD, and the results are used to obtain a description of the right singular matrix $\mathbf{V}$. To construct $\mathbf{W}$, we sample from a categorical distribution that takes $i \in$

$\{1, 2, \cdots, I\}$ with probability

$$\mathcal{F}(i) = \frac{\|\mathbf{X}(i,:)\|^2}{\|\mathbf{X}\|_F^2}. \tag{15}$$

Following the convention, we denote this categorical distribution by $\mathcal{F}$. As a result of sampling from this categorical distribution $P$ times, we obtain indices $i_1, i_2, \cdots, i_P$ specifying the rows in $\mathbf{X}$. The columns of $\mathbf{W}$ are then defined by sampling from a categorical distribution that takes $j \in \{1, 2, \cdots, J\}$ with probability

$$\mathcal{G}(j) = \frac{1}{P} \sum_{p=1}^{P} \frac{\mathbf{X}(i_p, j)^2}{\|\mathbf{X}(i_p,:)\|^2}. \tag{16}$$

Here, we denote this categorical distribution by $\mathcal{G}$. Sampling from this categorical distribution $P$ times, we obtain indices $j_1, j_2, \cdots, j_P$. The indices $\{i_p\}_{p=1}^{P}$ and $\{j_p\}_{p=1}^{P}$ are then used to construct the $P \times P$ matrix $\mathbf{W}$ whose $(p, q)$-th entry is

$$\frac{\mathbf{X}(i_p, j_q)}{P\sqrt{\mathcal{F}(i_p)\mathcal{G}(j_q)}}. \tag{17}$$

The above matrix construction is performed according to Algorithm 2.

---

**Algorithm 2: Matrix sampling**

**Input:** Input matrix $\mathbf{X} \in \mathbb{R}^{I \times J}$, parameter $P$

**Output:** Matrix $\mathbf{W} \in \mathbb{R}^{P \times P}$, set of indices $\{i_p\}_{p=1}^{P}$

1: For $p = 1$ to $P$ do
2:   Sample from the categorical distribution $\mathcal{F}$. Denote the result by $i_p$.
3: End for
4: For $p = 1$ to $P$ do
5:   Sample from the categorical distribution $\mathcal{G}$. Denote the result by $j_p$.
6: End for
7: Construct the $P \times P$ matrix $\mathbf{W}$ whose $(p, q)$-th entry is $\frac{\mathbf{X}(i_p, j_q)}{P\sqrt{\mathcal{F}(i_p)\mathcal{G}(j_q)}}$.

---

The qiSVD algorithm applies SVD to the above $P \times P$ matrix $\mathbf{W}$. By denoting the SVD results by $\mathbf{W} = \mathbf{U}_W \mathbf{\Sigma}_W \mathbf{V}_W^\mathrm{T}$, we produce a matrix $\mathbf{S} \in \mathbb{R}^{P \times J}$ and vectors $\mathbf{u}_1, \cdots, \mathbf{u}_K \in \mathbb{R}^P$ so that the vectors $\mathbf{S}^\mathrm{T} \mathbf{u}_1, \cdots, \mathbf{S}^\mathrm{T} \mathbf{u}_K$ span a space similar to that spanned by the first $K$ right singular vectors of the input matrix $\mathbf{X}$. According to the previous studies (Frieze et al., 1998; Tang, 2019), we set $\mathbf{S}$ as the $P \times J$ matrix whose $p$-th row is $\mathbf{X}(i_p,:)$, and $\mathbf{u}_k$ as the vector whose $p$-th element is

$$\frac{\mathbf{U}_W(p,k)}{\mathbf{\Sigma}_W(k,k)\sqrt{P\mathcal{F}(i_p)}} . \qquad (18)$$

The set $\{\mathbf{S}, \{\mathbf{u}_k\}_{k=1}^{K}\}$ forms a description of the right singular vectors of $\mathbf{X}$. The qiSVD algorithm is summarized in Algorithm 3.

---

**Algorithm 3: qiSVD**

**Input:** Input matrix $\mathbf{X} \in \mathbb{R}^{I \times J}$, parameters $K$, $P$

**Output:** Description $\{\mathbf{S}, \{\mathbf{u}_k\}_{k=1}^{K}\}$, set of indices $\{i_p\}_{p=1}^{P}$

1: $[\mathbf{W}, \{i_p\}_{p=1}^{P}] \leftarrow \mathrm{MatrixSampling}(\mathbf{X}, P)$

2: $[\mathbf{U}_W, \mathbf{\Sigma}_W, \mathbf{V}_W] \leftarrow \mathrm{SVD}(\mathbf{W})$

3: Let $\mathbf{S}$ be the $P \times J$ matrix whose $p$-th row is $\mathbf{X}(i_p,:)$.

4: Let $\hat{\mathbf{u}}_k \in \mathbb{R}^P$ $(k = 1, \cdots, K)$ be the vector whose $p$-th element is $\mathbf{U}_W(p,k)/\left(\mathbf{\Sigma}_W(k,k)\sqrt{P\mathcal{F}(i_p)}\right)$.

5: Orthonormalize $\{\mathbf{S}^\mathrm{T}\hat{\mathbf{u}}_k\}_{k=1}^{K}$ through the Gram–Schmidt process, and obtain a new set of vectors $\{\mathbf{u}_k\}_{k=1}^{K}$ so that $\{\mathbf{S}^\mathrm{T}\mathbf{u}_k\}_{k=1}^{K}$ is an orthonormal set (see **Appendix** for complete explanation).

---

The original qiSVD algorithm proposed by Tang (2019) excludes Step 5 of the above algorithm and produces an off-orthonormal set of vectors. The orthonormalization step (see Appendix for details) allows deriving an orthonormal set of vectors for qiCCA. Previous studies derived the theoretical upper bounds of the approximation error of qiSVD (Frieze et al., 1998; Tang, 2019), and experimentally tested its approximation performance on artificial data (Arrazola et al., 2019). In the present study, we experimentally demonstrate the approximation performance of qiSVD on both artificial and real datasets.

### 3.4. Time complexity

To execute Step 1 in Algorithm 3 on an input matrix $\mathbf{X} \in \mathbb{R}^{I \times J}$ in $O(\log(IJ))$ computational time, we adopt the data structure proposed in Tang (2019). The computational time of Steps 2–5 is independent of $I$. Therefore, qiSVD can be performed with a computational time proportional to the logarithm of $I$. Below, we explain how to execute Algorithm 2 (*i.e.*, Step 1 in Algorithm 3) in $O(\log(IJ))$ computational time.

Steps 2 and 5 in Algorithm 2 can be computed fast using binary search trees. For simplicity, we first suppose that an $N$-dimensional vector $\mathbf{x} \in \mathbb{R}^N$ is given, and we sample from a categorical distribution that takes $n$ $(n = 1,2,\cdots,N)$ with probability $\mathbf{x}(n)^2/\|\mathbf{x}\|^2$. We assume that the elements of $\mathbf{x}$ are stored in a binary search tree with $N$ leaves (Figure 1A). In this tree, the $n$-th leaf contains $\mathbf{x}(n)^2$ and each internal node contains the sum of its two child nodes. By executing the binary search algorithm, for a given value $u \in [0,1]$, we can find the index $n'$ satisfying

$$\frac{1}{\|\mathbf{x}\|^2} \sum_{n=1}^{n'-1} \mathbf{x}(n)^2 \leq u < \frac{1}{\|\mathbf{x}\|^2} \sum_{n=1}^{n'} \mathbf{x}(n)^2 \tag{19}$$

in $O(\log(N))$ computational time. Therefore, by sampling $u$ from the uniform distribution, we can sample from a categorical distribution that takes $n$ $(n = 1,2,\cdots,N)$ with probability $\mathbf{x}(n)^2/\|\mathbf{x}\|^2$ in $O(\log(N))$ computational time. This technique is faster than naïve sampling algorithms without binary search trees, such as, for example, the one adopted in SciPy (Figure 1B). To perform the sampling operations in Step 2 of Algorithm 2 on an input matrix $\mathbf{X} \in \mathbb{R}^{I \times J}$, we prepare a binary search tree with $I$ leaves, which performs the required samplings in $O(\log I)$ time. To perform the sampling operations in Step 5 of Algorithm 2, we prepare $I$ binary search trees with $J$ leaves storing the individual rows of the matrix. In Step 5, we uniformly and randomly select an index from $\{i_1, i_2, \cdots, i_p\}$, then sample from the categorical distribution using the binary search tree corresponding to the row specified by the selected index. This sampling process is equivalent to Step 5 of Algorithm 2 (*i.e.*, sampling from the distribution $\mathcal{G}$), and is completed in $O(\log J)$ computational time. A total of $(I +$

1) binary search trees are used to store the matrix. Variants of this binary tree data structure are described in the literature (Kerenidis & Prakash, 2016; Tang, 2019).

### 3.5. Quantum-inspired canonical correlation analysis

In qiCCA, to reduce the computational time required to run CCA, we replace $\mathbf{U}_1$ and $\mathbf{U}_2$ in Eq. (10) with their descriptions. To obtain descriptions of $\mathbf{U}_1$ and $\mathbf{U}_2$, we apply qiSVD to $\mathbf{X}^{\mathrm{T}}$ and $\mathbf{Y}^{\mathrm{T}}$, respectively. Here, the descriptions of $\mathbf{U}_1$ and $\mathbf{U}_2$ are denoted by $\{\mathbf{S}^{(1)}, \{\mathbf{u}^{(1)}_{l_1}\}_{l_1=1}^{L_1}\}$ and $\{\mathbf{S}^{(2)}, \{\mathbf{u}^{(2)}_{l_2}\}_{l_2=1}^{L_2}\}$, respectively. Using Eq. (14), the canonical variate $\mathbf{C}_X = \mathbf{X}\mathbf{W}_X$ is rewritten as $\mathbf{U}_1 \mathbf{U}_3(:,1:K)$. Replacing $\mathbf{U}_1$ with $\mathbf{S}^{(1)\mathrm{T}}[\mathbf{u}^{(1)}_1, \mathbf{u}^{(1)}_2, \cdots, \mathbf{u}^{(1)}_{L_1}]$, $\mathbf{X}\mathbf{W}_X$ is then approximated as follows:

$$\begin{aligned} \mathbf{X}\mathbf{W}_X &= \mathbf{U}_1 \mathbf{U}_3(:,1:K) \\ &\approx \mathbf{S}^{(1)\mathrm{T}}[\mathbf{u}^{(1)}_1, \cdots, \mathbf{u}^{(1)}_{L_1}] \mathbf{U}_3(:,1:K) \\ &= \mathbf{X}\mathbf{A}^{(1)\mathrm{T}}[\mathbf{w}^{(1)}_1, \cdots, \mathbf{w}^{(1)}_K], \end{aligned} \quad (20)$$

where $\mathbf{A}^{(1)}$ is the $P_1 \times D_1$ binary matrix such that

$$\mathbf{S}^{(1)\mathrm{T}} = \mathbf{X}\mathbf{A}^{(1)\mathrm{T}}, \quad (21)$$

and $\mathbf{w}^{(1)}_k \; (k = 1, \cdots, K)$ are the vectors determined as

$$[\mathbf{w}^{(1)}_1, \cdots, \mathbf{w}^{(1)}_K] = [\mathbf{u}^{(1)}_1, \cdots, \mathbf{u}^{(1)}_{L_1}] \mathbf{U}_3(:,1:K). \quad (22)$$

According to Eq. (20), $\mathbf{A}^{(1)\mathrm{T}}[\mathbf{w}^{(1)}_1, \cdots, \mathbf{w}^{(1)}_K]$ can be regarded as an approximation of $\mathbf{W}_X$. Therefore, the set $\{\mathbf{A}^{(1)}, \{\mathbf{w}^{(1)}_k\}_{k=1}^{K}\}$ can be regarded as a description of $\mathbf{W}_X$. A similar process is executed to approximate $\mathbf{W}_Y$. The descriptions of $\mathbf{W}_X$ and $\mathbf{W}_Y$ are computed using Algorithm 4.

> **Algorithm 4: qiCCA**
>
> **Input:** Pair of input matrices $\mathbf{X} \in \mathbb{R}^{N \times D_1}$, $\mathbf{Y} \in \mathbb{R}^{N \times D_2}$, parameters $K$, $L_1$, $L_2$, $P_1$, $P_2$
>
> **Output:** Pair of descriptions $\{\mathbf{A}^{(1)}, \{\mathbf{w}^{(1)}_{l_1}\}_{l_1=1}^{L_1}\}$, $\{\mathbf{A}^{(2)}, \{\mathbf{w}^{(2)}_{l_2}\}_{l_2=1}^{L_2}\}$
>
> 1: $\left[\mathbf{S}^{(1)}, \{\mathbf{u}^{(1)}_{l_1}\}_{l_1=1}^{L_1}, \{i^{(1)}_{p_1}\}_{p_1=1}^{P_1}\right] \leftarrow \text{qiSVD}(\mathbf{X}^\mathrm{T}, L_1, P_1)$
>
> 2: $\left[\mathbf{S}^{(2)}, \{\mathbf{u}^{(2)}_{l_2}\}_{l_2=1}^{L_2}, \{i^{(2)}_{p_2}\}_{p_2=1}^{P_2}\right] \leftarrow \text{qiSVD}(\mathbf{Y}^\mathrm{T}, L_2, P_2)$
>
> 3: $[\mathbf{U}_3, \mathbf{\Sigma}_3, \mathbf{V}_3] \leftarrow \text{SVD}\left(\left[\mathbf{u}^{(1)}_1, \mathbf{u}^{(1)}_2, \cdots, \mathbf{u}^{(1)}_{L_1}\right]^\mathrm{T} \mathbf{S}^{(1)} \mathbf{S}^{(2)\mathrm{T}} \left[\mathbf{u}^{(2)}_1, \mathbf{u}^{(2)}_2, \cdots, \mathbf{u}^{(2)}_{L_2}\right]\right)$
>
> 4: $\left[\mathbf{w}^{(1)}_1, \cdots, \mathbf{w}^{(1)}_K\right] \leftarrow \left[\mathbf{u}^{(1)}_1, \cdots, \mathbf{u}^{(1)}_{L_1}\right] \mathbf{U}_3(:, 1:K)$
>
> 5: $\left[\mathbf{w}^{(2)}_1, \cdots, \mathbf{w}^{(2)}_K\right] \leftarrow \left[\mathbf{u}^{(2)}_1, \cdots, \mathbf{u}^{(2)}_{L_2}\right] \mathbf{V}_3(:, 1:K)$
>
> 6: Let $\mathbf{A}^{(1)}$ be the $P_1 \times D_1$ binary matrix whose $(p, d)$-th entry is
> $$\begin{cases} 1 & (i^{(1)}_p = d) \\ 0 & (\text{otherwise}) \end{cases}$$
>
> 7: Let $\mathbf{A}^{(2)}$ be the $P_2 \times D_2$ binary matrix whose $(p, d)$-th entry is
> $$\begin{cases} 1 & (i^{(2)}_p = d) \\ 0 & (\text{otherwise}) \end{cases}$$

To apply qiSVD to $\mathbf{X}^\mathrm{T}$ and $\mathbf{Y}^\mathrm{T}$, we assume that both input matrices are stored as binary tree data (see Subsection 3.4). Therefore, qiSVD using binary search trees can be used to execute qiCCA in $O(\log(D_1 D_2))$ computational time.

The parameters $L_1$, $L_2$, $P_1$, and $P_2$ influence the performance of qiCCA. For simplicity, we set $L_1$ and $L_2$ to the same value $L$, and $P_1$ and $P_2$ to the same value $P$. According to previous qiSVD studies (Arrazola et al., 2019; Tang, 2019), increasing $P$ results in improving the performance of the algorithm but augments the computational time. In Subsection 4.5, we evaluate the effect of $P$ on the performance of qiSVD using real datasets. Based on the results obtained in Subsection 4.5, in Subsection 4.6, we set $P = 1.5L$ in qiCCA, and evaluate the effect of $L$ on the performance of qiCCA. Unless stated otherwise, we set $L = 0.5 \max(D_1, D_2)$. We also demonstrate that the orthonormalization step in our modified qiSVD significantly improves the performance of qiCCA. Concerning the dependency of the computational time on $P$, both qiSVD and qiCCA require $O(P^3)$ computational time, as SVD is applied to a $P \times P$ matrix in qiSVD. In qiCCA, SVD is also applied to an $L \times L$ matrix. Therefore, the computational time of qiCCA with respect to $L$ is $O(L^3)$. In the case of using the above parameter setting (*i.e.*, $P = 1.5L$ and $L = 0.5\max(D_1, D_2)$), the computational time of qiCCA with respect to the input dimensionalities $D_1$ and $D_2$ is $O(\max(D_1, D_2)^3)$. When these parameters are manually set to constant values, the computational time with respect to the input dimensionalities is $O(\log D_1 + \log D_2)$.

# 4. Experiments

In this section, we describe the experimental setup and the results of the conducted experiments. In Subsections 4.1, 4.2, and 4.3, we describe synthetic data, real data, and evaluation metrics used in the conducted experiments, respectively. Subsection 4.4 explains variants of the CCA algorithm considered in this study for the purpose of comparison. Subsections 4.5–4.7 outline the experimental results. All experiments were conducted using an Intel CPU Xeon Gold 5115 (2.4 GHz, 768 GB memory).

## 4.1. Synthetic datasets

Experiments were conducted on two types of synthetic data. Using the first dataset, we compared the performance of qiSVD and the conventional SVD. The second dataset was employed to compare the performances of qiCCA and the conventional CCA.

To compare the performances of qiSVD and the conventional SVD, we synthesized an input matrix $\mathbf{X} \in \mathbb{R}^{I \times J}$ as $\mathbf{X} = \mathbf{ZB}$, where $\mathbf{Z}$ and $\mathbf{B}$ were $(I \times 100)$- and $(100 \times J)$-matrices composed of entries sampled from the standard normal distribution. Fixing $I$ at 10000, we varied $J$ across $\{2^5, 2^6, \cdots 2^{15}\}$, and examined the dependency of the computational time on the input dimensionality.

To compare qiCCA and the conventional CCA, we prepared a pair of input matrices $\mathbf{X} \in \mathbb{R}^{N \times D_1}$ and $\mathbf{Y} \in \mathbb{R}^{N \times D_2}$ following the standard assumption of probabilistic CCA (Bach & Jordan, 2005) formulated as follows:

$$\mathbf{X} = \mathbf{ZB}_1 + 0.5\mathbf{E}_1, \quad \mathbf{Y} = \mathbf{ZB}_2 + 0.5\mathbf{E}_2, \quad (23)$$

where $\mathbf{Z} \in \mathbb{R}^{N \times K}$, $\mathbf{B}_1 \in \mathbb{R}^{K \times D_1}$, $\mathbf{B}_2 \in \mathbb{R}^{K \times D_2}$, $\mathbf{E}_1 \in \mathbb{R}^{N \times D_1}$, and $\mathbf{E}_2 \in \mathbb{R}^{N \times D_2}$ are matrices composed of entries sampled from the standard normal distribution. Here, *N* and *K* were set to 10000 and 100, respectively, and $D_1$ and $D_2$ were set to the same value, which varied across $\{2^5, 2^6, \cdots 2^{15}\}$. Using $\mathbf{X}$ and $\mathbf{Y}$ as input, we examined the dependency of the computational time on the input dimensionality.

## 4.2. Real datasets

The computational times and performances of qiSVD and qiCCA were tested on five real datasets, each with two views. While evaluating CCA and its variants, the data of the two views were processed as a pair of input datasets. In the case of evaluating SVD and qiSVD, we used only the data of the first view. The five datasets are described below.

*4.2.1. Real dataset 1: MNIST*

The first real dataset was the MNIST dataset (Lecun et al., 1998), which was employed as a benchmark dataset in previous tests of CCA variants (Andrew et al., 2013; W. Wang et al., 2015). According to the previously introduced procedure (Andrew et al., 2013), we treated the pixel values of the 14 left and 14 right columns in each $28 \times 28$ handwritten digit image as one pair of input data. The number of input dimensions was 392 per view. Following the dataset specifications, we allocated 50000, 10000 and 10000 samples to the training, validation and test datasets, respectively.

*4.2.2. Real dataset 2: CIFAR10*

The second real dataset was the CIFAR10 image dataset (Krizhevsky, 2009). We converted the color images into grayscale ones, then (similarly to the MNIST data) treated the pixel values of the 16 left and 16 right columns in each $32 \times 32$ object image as one pair of input data. The number of input dimensions was 512 per view. The original dataset contains 50000 training samples and 10000 test samples. As the validation data, we randomly selected 10000 samples from the 50000 training samples, and utilized the remaining 40000 samples as the training data.

*4.2.3. Real dataset 3: WikiCLIR*

The third real dataset was the WikiCLIR dataset (Schamoni et al., 2014) used for cross-lingual (German–English) information retrieval. It contains pairs of German and English sentences representing the same contents. In the present study, the German and English sentences were respectively transformed into 300-dimensional vectors by Wikipedia2Vec (Yamada et al., 2018). The German and English data were treated as view 1 and view 2, respectively. Following the dataset specifications, we allocated 50000, 10000 and 10000 samples to the training, validation and test data, respectively.

*4.2.4. Real dataset 4: Japanese–English Subtitle Corpus (JESC)*

The fourth real dataset was the Japanese–English Subtitle Corpus (JESC) dataset (Pryzant et al., 2017), which contains pairs of Japanese and English sentences representing the same contents. Similarly as in the case of the WikiCLIR dataset, the Japanese and English sentences were respectively transformed into 300-dimensional vectors by Wikipedia2Vec (Yamada et al., 2018). The Japanese and English data were treated as view 1 and view 2, respectively. Following the dataset specifications, we allocated 50000, 2000 and 2000 samples to the training, validation and test data, respectively.

*4.2.5. Real dataset 5: X-ray Microbeam (XRMB) Database*

The fifth real dataset was the X-ray Microbeam (XRMB) database (Westbury, 1994), which contains pairs of simultaneously recorded acoustic and articulatory data obtained from speeches pronounced by subjects. The acoustic vector (with 273 dimensions) and articulatory vector (with 112 dimensions) were treated as view 1 and view 2, respectively. According to the dataset specifications, we allocated 25873, 8624 and 8625 samples to the training, validation and test datasets, respectively.

**4.3. Evaluation metrics**

To evaluate the performances of qiSVD and the conventional SVD, we computed the following metric:

$$1 - \frac{\|\mathbf{X} - \mathbf{X}\mathbf{V}\mathbf{V}^\mathrm{T}\|_F^2}{\|\mathbf{X}\|_F^2}, \qquad (24)$$

where $\mathbf{X} \in \mathbb{R}^{I \times J}$ is the input matrix and $\mathbf{V} \in \mathbb{R}^{J \times K}$ is the matrix composed of the (exact or approximate) right singular vectors corresponding to the top $K$ singular values. The term $\|\mathbf{X} - \mathbf{X}\mathbf{V}\mathbf{V}^\mathrm{T}\|_F^2$, which measures the difference between the input matrix $\mathbf{X}$ and its low-rank approximation $\mathbf{X}\mathbf{V}\mathbf{V}^\mathrm{T}$, is frequently used as the objective function in low-rank matrix approximation algorithms (Achlioptas & Mcsherry, 2007).

The performances of qiCCA and the CCA variants were evaluated by two metrics, the "sum of correlations" and "area under the ROC curve." The sum of correlations was computed according to a previously introduced procedure (Andrew et al., 2013). Specifically, we computed the correlation coefficient for each pair of canonical variates, and summed them across the first $K$ pairs. Here, $K$ was ranged from 1 to 100. While comparing qiCCA and the conventional CCA, we focused on evaluating the similarity between the two algorithms rather than on estimating the generalization performance. For this purpose, the sums of the correlations were computed using the training data. In this case, the conventional CCA always outperformed qiCCA, and we investigated the extent of the performance degradation of qiCCA relative to the conventional CCA. In contrast, while comparing qiCCA and other variants of CCA, we aimed at estimating the generalization performance. Accordingly, the sums of the correlations were computed using the test data. To evaluate the statistical significance of the difference in this metric, we performed a statistical test. The given test data were randomly split into 100 groups, and the sum of the correlations obtained by each CCA variant was computed using each of the 100 groups. Assuming the resulting 100 values as 100 samples, we applied the signed-rank test to compare the median of this metric between a pair of CCA variants.

The second performance metric, the area under the ROC curve, was computed as described in Chu et al. (2013). This metric was computed through item retrieval using the canonical variates obtained by CCA. Specifically, we computed

$$\tau(n) = \|\mathbf{C}_X(n_{\text{target}},:) - \mathbf{C}_Y(n,:)\|, \tag{25}$$

for $n = 1, \cdots, N$, where $\mathbf{C}_X \in \mathbb{R}^{N \times K}$ and $\mathbf{C}_Y \in \mathbb{R}^{N \times K}$ are the first $K$ pairs of the canonical variates, and $n_{\text{target}} \in \{1, \cdots, N\}$ is the sample number (*i.e.*, item number) to be retrieved. After sorting $\tau(1), \cdots, \tau(N)$ in descending order, we computed

$$1 - \frac{\#\{n \in \{1, \cdots, N\} \mid \tau(n) < \tau(n_{\text{target}})\}}{N}, \tag{26}$$

where $\#\{\cdot\}$ denotes the number of the members in the set $\{\cdot\}$. Conventionally, this value is called the area under the ROC curve. We computed Eq. (26) for $n_{\text{target}} = 1, \cdots, N$ and averaged the results across the $N$ samples. In the following section, we report the resulting mean area under the ROC curve. To evaluate the generalization performance, this value was computed using the test data.

### 4.4. Other variants of CCA
We also compared the performances of qiCCA with those of two linear and four nonlinear

CCA variants. The linear variants were scalable CCA (Ma et al., 2015) and sparse CCA (Chu et al., 2013; Hardoon & Shawe-Taylor, 2011; Witten et al., 2009). Among several sparse CCA formulations, we implemented the algorithm proposed by Chu et al. (2013). The nonlinear variants were kernel CCA (Akaho, 2001; Bach & Jordan, 2002; Hardoon et al., 2004; Melzer et al., 2001), approximate kernel CCA (W. Wang & Livescu, 2016), deep CCA (Andrew et al., 2013), and deep canonically correlated autoencoders (DCCAE) (W. Wang et al., 2015). The regularization coefficients of sparse CCA, scalable CCA, kernel CCA, and approximate kernel CCA were optimized in performance evaluation on the validation data. In kernel CCA and approximate kernel CCA, we employed a second-order polynomial kernel and a Gaussian kernel, respectively. It should be noted that approximate kernel CCA can be combined only with shift-invariant kernels (*e.g.*, Gaussian kernels) and is incompatible with polynomial kernels. The parameters of the kernel functions were optimized using the validation data. In all considered algorithms, the other hyperparameters and architectures were set according to the default settings reported in the original studies.

## 4.5. Evaluation of qiSVD

First, we compared the computational times of qiSVD and the conventional SVD on the synthetic data, namely, on a $10000 \times J$ real matrix **X** (see Subsection 4.1). Here, the number of dimensions $J$ was varied (see Figure 2A), and the qiSVD parameters $K$ and $P$ were set to 100 and 150, respectively. The computational times of both algorithms were similar for $J$ less than $2^{11}$; however, qiSVD was exponentially faster than the conventional SVD with an increase in $J$. We also examined the computational time required to complete Step 1 in qiSVD (*i.e.*, matrix sampling). In the case of $J$ varying from $2^5$ to $2^{15}$, Step 1 required 15%–20% of the total computational time.

Next, we compared the performances of the two algorithms on the same synthetic data. The performance metric was $1 - (\|\mathbf{X} - \mathbf{XVV}^\mathrm{T}\|_F^2 / \|\mathbf{X}\|_F^2)$ where **V** is the $J \times K$ (exact or approximate) right singular matrix (see Subsection 4.3). This value measures the recovery extent of the input matrix **X** by its low-rank approximation $\mathbf{XVV}^\mathrm{T}$. The number of columns in the input matrix (*i.e.*, $J$) was fixed at 10000 while $K$ was varied (see Figure 2B). The conventional SVD outperformed qiSVD by only several percent, indicating that the subspace spanned by the vectors generated by qiSVD was similar to that spanned by the exact right singular vectors.

The above analysis was repeated on the five real datasets (see Subsection 4.2). In these experiments, we set $K$ to 100 and the qiSVD parameter $P$ to 150. On all five datasets, the computational time was much smaller in the case of using qiSVD compared with that of the conventional SVD (Figure 3A), albeit with small performance losses (1.9%–6.1% lower than in the conventional SVD) (Figure 3B).

Next, we examined the effect of the parameter $P$ on the performance of qiSVD. Figure 4 illustrates the relative performance of qiSVD (the performance of qiSVD divided by that of the conventional SVD) as a function of $P$. As in the previous experiment, $K$ was set to 100. The performance of qiSVD increased with an increase in $P$. In all subsequent experiments, the parameter $P$ in qiSVD ($P_1$ and $P_2$ in qiCCA) was set to $1.5K$.

## 4.6. Evaluation of qiCCA

Then, we compared the computational times of qiCCA and the conventional CCA on the synthetic datasets. For this purpose, we prepared a pair of input matrices $\mathbf{X} \in \mathbb{R}^{N \times D_1}$ and $\mathbf{Y} \in \mathbb{R}^{N \times D_2}$ (see Subsection 4.1), and set $N$ to 10000. Here, the qiCCA parameters $L_1$ and $L_2$ were set to 100. The input dimensionalities $D_1$ and $D_2$ were set to the same value and varied across $\{2^5, 2^6, \cdots 2^{15}\}$. Figure 5A represents the computational times of qiCCA and the conventional CCA as functions of $D_1$ (or $D_2$). For input dimensionalities smaller than $2^{11}$, the computational times of qiCCA and the conventional CCA were similar; in the case of large input dimensionalities, qiCCA demonstrated an exponential speed-up over the conventional CCA. The performances were evaluated as described in previous studies (Andrew et al., 2013; W. Wang et al., 2015), namely, the canonical correlation was calculated for each pair of canonical variates and the correlation coefficients were summed across the first $K$ canonical components (see Subsection 4.3) (Figure 5B). Here, $D_1$ and $D_2$ were set to 10000. The correlation extraction degrees of qiCCA and the conventional CCA were similar.

We also compared qiCCA and the conventional CCA on the five real datasets used in the evaluation of qiSVD (see Subsection 4.2). The input dimensionalities of the individual views in the five datasets ranged from 112 to 512. The qiCCA parameters $L_1$ and $L_2$ were set to the same value $L = 0.5 \max(D_1, D_2)$ where $D_1$ and $D_2$ are the input dimensionalities in views 1 and 2, respectively. The first 100 canonical components were derived. On all datasets, the computational time was considerably less in the case of using qiCCA compared with that of the conventional CCA (Figure 6A). Comparing the sums of correlations between qiCCA and the conventional CCA, calculated for the first 100 canonical components (Figure 6B), the difference in performance varied from 7% to 31% over the five datasets.

We then examined the effect of $L$ on the performance of qiCCA. For this purpose, $L$ was varied from $0.1 \max(D_1, D_2)$ to $1.0 \max(D_1, D_2)$. Figure 7 illustrates the performance of qiCCA (summed over the first 100 canonical correlations) as a function of $L$. Increasing $L$ resulted in improving the performance of qiCCA.

Finally, we examined the improvement resulting from our qiSVD modification. In the original qiSVD algorithm proposed by Tang (2019), the approximate right singular vectors are not orthonormal, while our modified qiSVD returns a set of orthonormal vectors. Using the same procedure as in Figure 6B, we compared the performances of the qiCCA algorithms based on the original and our modified qiSVD (Figure 8). The results suggest that our modification is critical for high-performance qiCCA.

### 4.7. Nonlinear correlation extraction by qiCCA

In this subsection, we demonstrate that the high computational efficiency of qiCCA facilitates correlation extraction when combined with nonlinear mapping into high-dimensional spaces. For each input matrix (*i.e.*, each view), we concatenated the raw input features and the products of the input features for all possible feature pairs (Majima et al., 2014). Then, we processed the resulting high-dimensional data using qiCCA. We refer to this method as "qiCCA + 2nd-order." Meanwhile, the method that applies qiCCA to the raw input features only is called "qiCCA + 1st-order." It should be noted that this procedure results in increasing the input dimensionality from $D$ to $D + 0.5D(D - 1)$ (where $D$ is the number of raw input features). The computational time of processing such high dimensionality is often intractable for the conventional CCA. In the qiCCA + 2nd-order method, the qiCCA parameters $L_1$ and $L_2$ were both set to 3000.

Then, we compared the performances of the above qiCCA methods and other linear and nonlinear variants of CCA (scalable CCA, sparse CCA, kernel CCA, approximate kernel CCA, deep CCA, DCCAE; see Subsection 4.4). The regularization coefficients and parameters of the kernels in these algorithms were optimized using the training and validation data in each dataset. When evaluating the computational time, the time of hyperparameter optimization was excluded.

The comparison of the computational times of the considered eight algorithms on the five real datasets is illustrated in Figure 9A. The qiCCA + 2nd-order method was capable of processing the high-dimensional 2nd-order features within a moderate computational time. It should be noted that, as preliminary analysis, we applied other linear CCA algorithms (*e.g.*, conventional CCA and sparse CCA) to the same high-dimensional 2nd-order features and observed that their computations were unfeasible due to out-of-memory error in the same computational environment.

To evaluate the generalization performances, we then summed the top 100 canonical correlations obtained on the independent test data (see Subsection 4.3). On datasets 1–3 and dataset 5, qiCCA + 2nd-order resulted in extracting more correlations compared with qiCCA + 1st-order (Figure 9B)($p < 10^{-17}$ in the signed-rank test; see Subsection 4.3). Moreover, on dataset 3, the performance of qiCCA + 2nd-order was the highest among the eight considered algorithms ($p < 10^{-14}$ in the signed-rank test; see Subsection 4.3). Also, by analyzing the area under the ROC curve, which is a performance metric frequently used in item retrieval (see Subsection 4.3), we confirmed that qiCCA + 2nd-order performed as well as or better than the other algorithms on datasets 1–3 (Figure 9C).

# 5. Discussion

In the present study, we first experimentally demonstrated the computational efficiency and approximation performance of qiSVD using synthetic and real datasets (Figures 2 and 3). Then, we introduced qiSVD into the CCA algorithm, which exponentially reduces the computational time of CCA, and demonstrated the computational efficiency and performance of our proposed qiCCA (Figures 5 and 6). The fast computation of qiCCA enables the direct application of CCA even after nonlinearly mapping the raw input data into high-dimensional spaces. To illustrate this capability, we used the second-order monomials and applied qiCCA to the resulting high-dimensional data. In terms of correlation extraction, qiCCA extracted more correlations than linear CCA and achieved comparable performance to state-of-the-art nonlinear variants of CCA (Figure 9).

To exploit the fast computation of qiCCA in nonlinear correlation extraction, we computed the products of the input features for all possible feature pairs, and applied qiCCA to the resulting high-dimensional data. As this procedure highly increases the input dimensionalities, it is often infeasible when using the conventional CCA. Although the kernel trick can be applied to process such nonlinear transformations implicitly, our approach based on explicit mappings has several advantages. Kernel CCA computes the kernel matrices for the training data, and then, implicitly executes CCA in high-dimensional spaces with a computational time that is independent of the input dimensionalities. However, solving the eigenvalue problem with $N \times N$-sized kernel matrices (where $N$ is the number of training samples) requires $O(N^3)$ computational time. Therefore, kernel CCA is inapplicable to datasets with a large number of samples. In contrast, qiCCA operates in $O(N)$ computational time. Although several existing algorithms can approximate kernel CCA in $O(N)$ computational time by using random Fourier features (Uurtio et al., 2019; W. Wang & Livescu, 2016), their applications are restricted to shift-invariant kernels, and cannot process polynomial kernels. The proposed combination of qiCCA and nonlinear mapping using a large number of basis functions provides an alternative approach for revealing nonlinear correlations.

## 6. Conclusions

In the present study, we developed an algorithm that approximates CCA with computational time proportional to the logarithm of the input dimensionality. The proposed quantum-inspired CCA (qiCCA) algorithm is based on a recently developed randomized SVD algorithm referred to as quantum-inspired SVD (qiSVD). The computational efficiencies and the approximation performances of qiSVD and qiCCA were evaluated on one synthetic and five real datasets. Owing to the fast computation of qiCCA, the proposed method was capable of executing CCA even after nonlinearly mapping raw input data into high-dimensional spaces. In the experimental evaluations, the combined qiCCA and nonlinear mapping using a large number of basis functions extracted more correlations than linear CCA, and demonstrated comparable performance to state-of-the-art nonlinear variants of CCA. Our results confirmed the utility of qiCCA, and suggested that quantum-inspired computation has the potential to unlock a new field in which exponentially large dimensional data can be analyzed.

# Appendix

---
Algorithm 5: Orthonormalization

**Input:** Description $\{\mathbf{S} \in \mathbb{R}^{P \times J}, \{\hat{\mathbf{u}}_k \in \mathbb{R}^P\}_{k=1}^K\}$

**Output:** Description $\{\mathbf{S}, \{\mathbf{u}_k\}_{k=1}^K\}$

1: Let $\mathbf{M}$ be the $P \times P$ matrix $\mathbf{S}\mathbf{S}^\mathrm{T}$

2: Let $\mathbf{u}_1$ be the $P$-dimensional vector

$$\frac{\hat{\mathbf{u}}_1}{\sqrt{\hat{\mathbf{u}}_1^\mathrm{T} \mathbf{M} \hat{\mathbf{u}}_1}}$$

3: For $k = 2$ to $K$ do

4:     Let $\mathbf{v}_k$ be the $P$-dimensional vector

$$\hat{\mathbf{u}}_k - \sum_{l=1}^{k-1} \left(\hat{\mathbf{u}}_k^\mathrm{T} \mathbf{M} \mathbf{u}_l\right) \mathbf{u}_l$$

5:     Let $\mathbf{u}_k$ be the $P$-dimensional vector

$$\frac{\mathbf{v}_k}{\sqrt{\mathbf{v}_k^\mathrm{T} \mathbf{M} \mathbf{v}_k}}$$

6: End for

---

Because Step 1 in Algorithm 5 consumes $O(J)$ computational time, Algorithm 5 also operates in $O(J)$ time.

# Acknowledgments

The authors express their gratitude to Chikako Koide for preparing the environment for the analysis.

# Funding

This research was supported by JSPS KAKENHI Grant number 20K16465.

# Competing interests

All authors declare no competing financial interests.

# Figures and figure captions

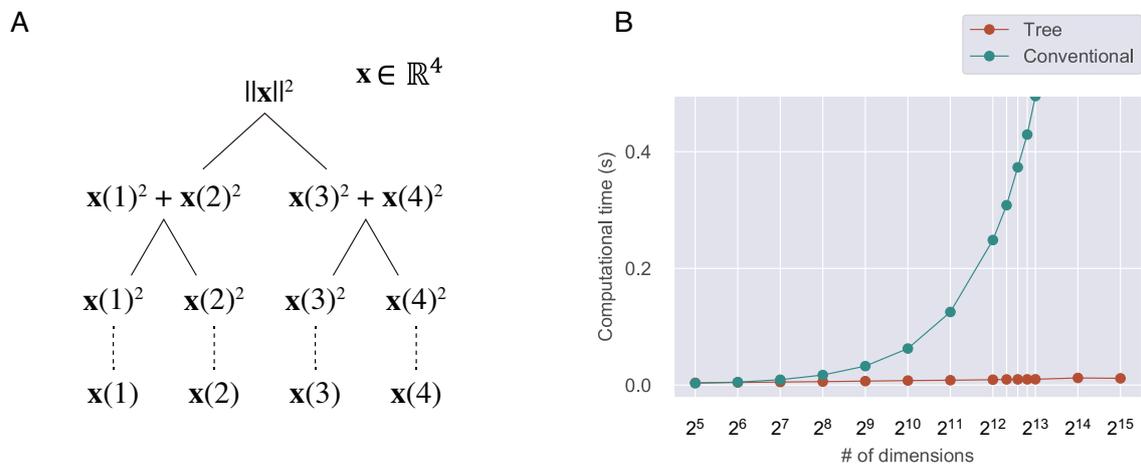

**Figure 1. Binary tree data structure and sampling. (A)** Binary tree data structure. An *N*-dimensional vector is stored as a binary search tree with *N* leaves. This allows sampling from a categorical distribution that takes *n* (*n* = 1, 2,…, *N*) with probability proportional to the square of the *n*-th element of the vector. **(B)** Computational time. The mean computational time to sample from the categorical distribution across ten repetitions is plotted as a function of the number of the dimensions. The algorithm using the binary tree data structure is compared to a naïve algorithm that takes time proportional to the number of the dimensions.

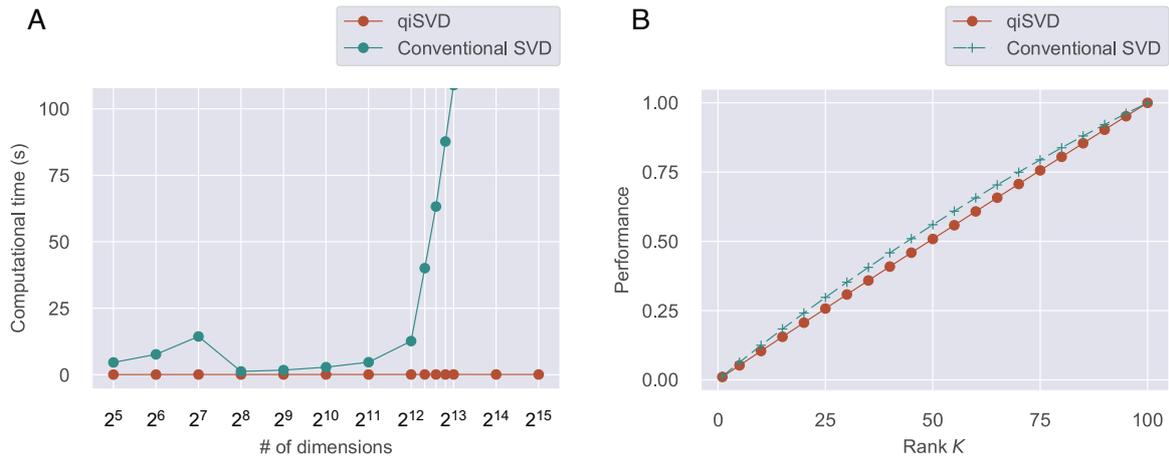

**Figure 2. Computational time and performance of quantum-inspired singular value decomposition (qiSVD) on synthetic datasets. (A)** Computational time evaluation. The mean computational times of qiSVD and the conventional SVD across ten repetitions are plotted as functions of the number of the columns of the input matrix. Here, the parameters $K$ and $P$ in qiSVD are set to 100 and 150. **(B)** Performance evaluation. The mean approximation performances of qiSVD and the conventional SVD across ten repetitions are plotted as functions of $K$. This plot illustrates how accurately a 10000 x 10000 input matrix is approximated by the rank-$K$ approximations obtained by qiSVD and the conventional SVD. The parameter $P$ in qiSVD is set to 150.

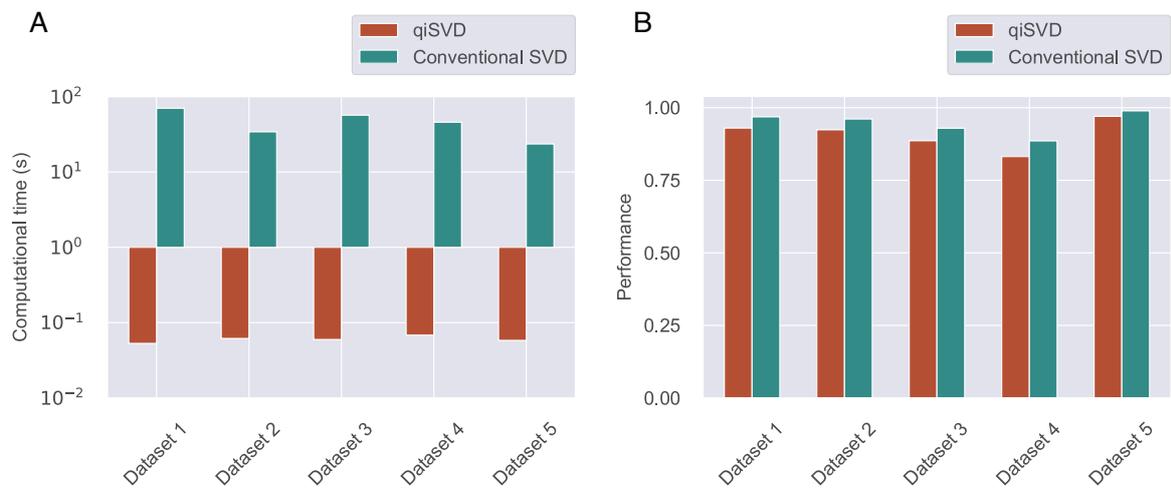

**Figure 3. Computational time and performance of qiSVD on real datasets. (A)** Computational time evaluation. Comparison of computational times of qiSVD and the conventional SVD on five real datasets. **(B)** Performance evaluation. The bar chart represents how accurately an input matrix is approximated by the rank-*K* approximations obtained by qiSVD and the conventional SVD. Here, *K* is set to 100 and the approximation performances are compared on the five real datasets.

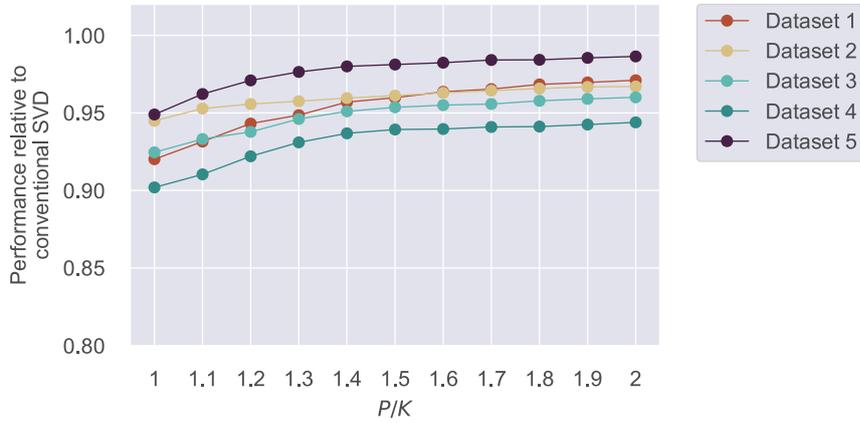

**Figure 4. Dependency of qiSVD performance on the parameter *P*.** This plot illustrates how accurately an input matrix is approximated by the rank-*K* approximation obtained using qiSVD with different *P* (the number of random samplings) (here, *K* = 100). The relative performance of qiSVD (the qiSVD performance divided by the conventional SVD performance) is plotted as a function of *P*/*K*. The results on the five real datasets are represented.

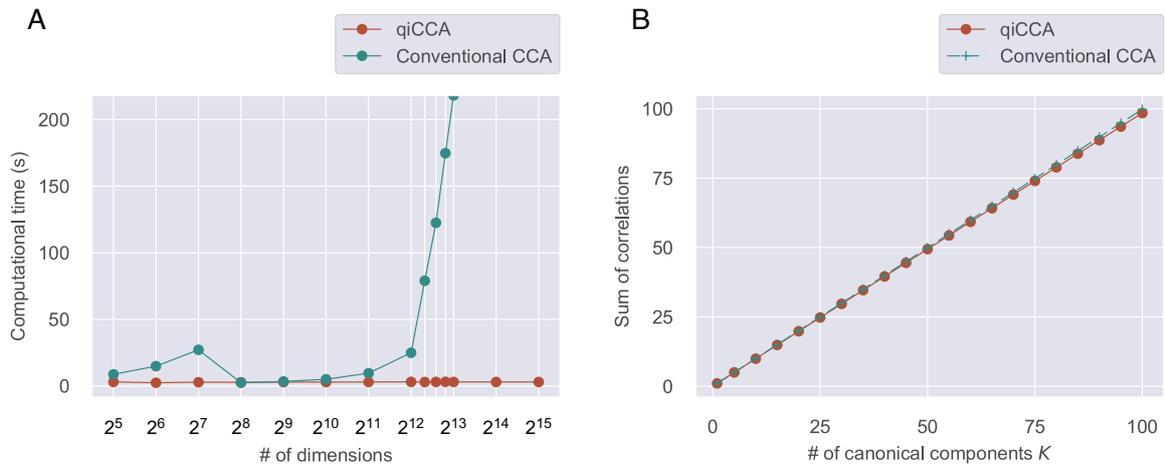

**Figure 5. Computational time and performance of quantum-inspired canonical correlation analysis (qiCCA) on synthetic datasets.** **(A)** Computational time evaluation. The mean computational times of qiCCA and the conventional CCA are averaged across ten repetitions, and the means are plotted as functions of the number of input dimensions (*i.e.,* input features). Here, the parameters $L$ and $P$ in qiCCA are set to 100 and 150. **(B)** Extractable correlations. The sums of the top $K$ canonical correlations in qiCCA and the conventional CCA are plotted as functions of $K$. Two 10000 x 10000 input matrices are used. Here, the parameters $L$ and $P$ in qiCCA are set to 100 and 150.

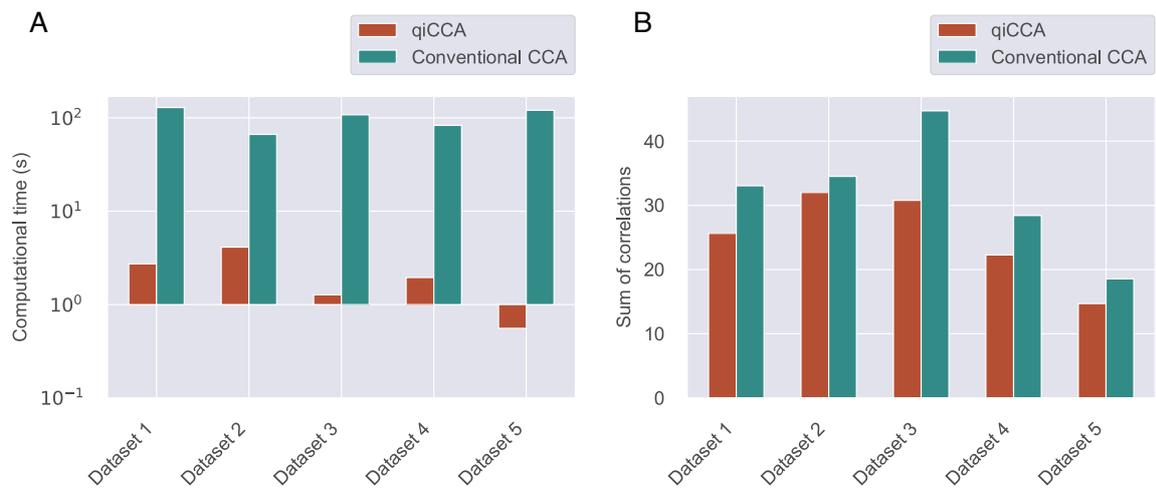

**Figure 6. Computational time and performance of qiCCA on real datasets. (A)** Computational time evaluation. Comparison of computational times of qiCCA and the conventional CCA on the five real datasets. **(B)** Extractable correlations. The sums of the top 100 canonical correlations in qiCCA and the conventional CCA. The results are compared on the five real datasets.

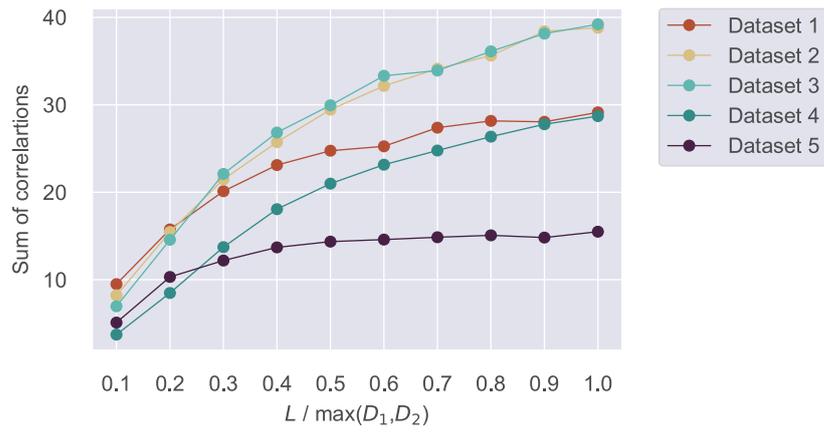

**Figure 7. Dependency of qiCCA performance on the parameter *L*.** The sums of the top 100 canonical correlations extracted by qiCCA versus the parameter *L*. The results on the five real datasets are represented.

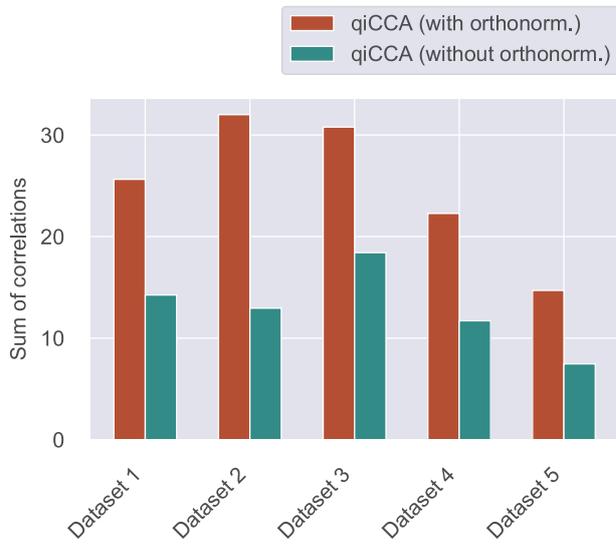

**Figure 8. Effect of the orthonormalization step.** To evaluate the performance improvement resulting from the orthonormalization step in qiCCA, this plot compares the performances of qiCCA with (red bars) and without (green bars) the orthonormalization step on the five real datasets.

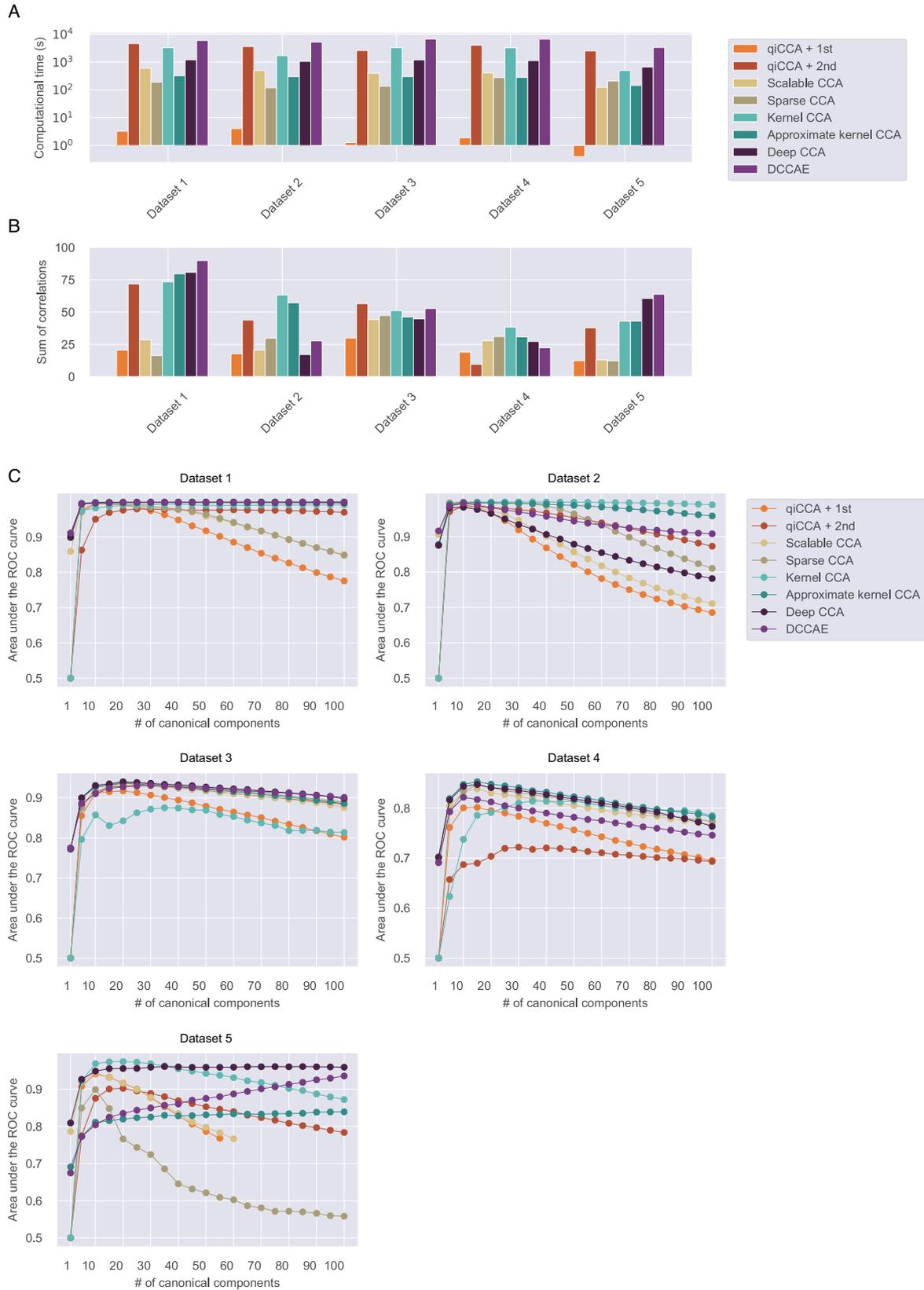

**Figure 9. Performance comparison of eight CCA variants. (A)** Computational time evaluation. The computational times of the eight CCA variants are compared on the five real datasets. **(B)** Extractable correlations. The sums of the top 100 canonical correlations

obtained by the eight CCA variants are compared on the five real datasets. **(C)** Item-retrieval performance. The item-retrieval performance of each CCA variant, evaluated by the area under the ROC curve, is plotted as a function of the number of used canonical components. The eight CCA variants are compared on each dataset. It should be noted that the numbers of extractable canonical components in qiCCA + 1st-order and scalable CCA are less than 100 on dataset 5 as the rank of the input matrix of view 2 is less than 100.